\documentclass[letterpaper, 10 pt, conference]{ieeeconf}  

\IEEEoverridecommandlockouts                              

\usepackage{graphicx}
\graphicspath{ {image/} }
\usepackage{amsmath, xparse}
\usepackage{subfigure}
\usepackage{multirow}
\usepackage{multicol}
\usepackage{makecell}
\usepackage{bm}
\usepackage[T1]{fontenc}
\usepackage{mathptmx} 
\usepackage{amsmath} 
\title{\LARGE \bf
Slippage-robust Gaze Tracking for Near-eye Display
}

\author{Wei Zhang$^{1}$, Jiaxi Cao$^{1}$, Xiang Wang$^{2}$, Enqi Tian$^{1}$ and Bin Li$^{1}$
\thanks{$^{1}$School of Information Science, 
        University of Science and Technology of China, HeFei 230026, China. {\tt\small 	zw1996@mail.ustc.edu.cn}}
\thanks{$^{2}$Department of Electrical and Computer Engineering, Carnegie Mellon University, Pittsburgh, USA. {\tt\small xiangw2@andrew.cmu.edu}}
\thanks{Corresponding author:Bin Li (binli@ustc.edu.cn)}%
}

\begin{document}

\maketitle
\thispagestyle{empty}
\pagestyle{empty}

\begin{abstract}
In recent years, head-mounted near-eye display devices have become the key hardware foundation for virtual reality and augmented reality. Thus head-mounted gaze tracking technology has received attention as an essential part of human-computer interaction. However, unavoidable slippage of head-mounted devices (HMD) often results higher gaze tracking errors and hinders the practical usage of HMD. To tackle this problem, we propose a slippage-robust gaze tracking for near-eye display method based on the aspheric eyeball model and accurately compute the eyeball optical axis and rotation center. We tested several methods on datasets with slippage and the experimental results show that the proposed method significantly outperforms the previous method (almost double the suboptimal method). 

\end{abstract}

\section{INTRODUCTION}

With the prosperity of the metaverse, the demand for augmented, virtual and mixed reality (AR, VR, and MR) is rapidly increasing. Head-mounted near-eye display, functioning as the hardware foundation of all three techniques,  project virtual scenes onto the human eye and create an immersive environment. Although various interaction methods, including gripping and keyboard, are explored~\cite{liu2019high,lin2022comparison}, eye gaze movement is still the most natural interactive way in HMD. Thus, head-mounted near-eye gaze tracking technology shows excellent potential in human-computer interaction. 

Based on knowledge of human gaze mode in the fundamental environment, gaze-based improvements in graphic rendering and displaying~\cite{patney2016towards,konrad2020gaze,krajancich2020optimizing} have been studied. Furthermore, the human gaze directly indicates human perceptions and attention, which is suitable for human-computer interaction. Gaze cues have already been adopted in interactions between human and intelligent systems~\cite{zheng2018rapid,majaranta2019eye,saran2018human} and similar interaction studies have been conducted on AR/VR devices with gaze tracking implementation. Results show that gaze-based interaction promises higher interaction quality, speed, and a more user-friendly experience~\cite{blattgerste2018advantages}. Unfortunately, in common AR/VR application scenarios, body motions and head motions constantly exist during practical use, which leads to an unavoidable HMD slippage and slight slippage often results in a significant increase in gaze estimation error of head-mounted devices~\cite{li2020optical,clay2019eye}. Most commercial eye trackers also suffer from the accuracy degradation caused by slippage~\cite{niehorster2020impact}. Methods to maintain gaze accuracy after slippage have been investigated~\cite{santini2019get} and results show the robustness of headset slippage in head-mounted gaze tracking. However, the overall gaze estimation accuracy of slippage-robust methods is lower than most state-of-the-art methods and only qualified for limited uses.  
 
In this work, we proposed an eyeball optical axis and position estimation method based on the aspheric eye model. Then we proposed a gaze tracking geometric model for the near-eye display to estimate scene gaze points. Specially, we set up a low-cost hardware device to validate our slippage robust gaze tracking method for near-eye display. We conducted experiments on nine subjects and asked them to remount and rotate the device repeatedly to simulate slippage in actual use. The experiment results show that our method outperforms the state-of-art method \cite{santini2019get} by 100\%, the angular offset decreases from $1.51^\circ$ to $0.76^\circ$.

\section{RELATED WORK}



\begin{figure}[ht]
    \centering
    \includegraphics[width=0.8\linewidth]{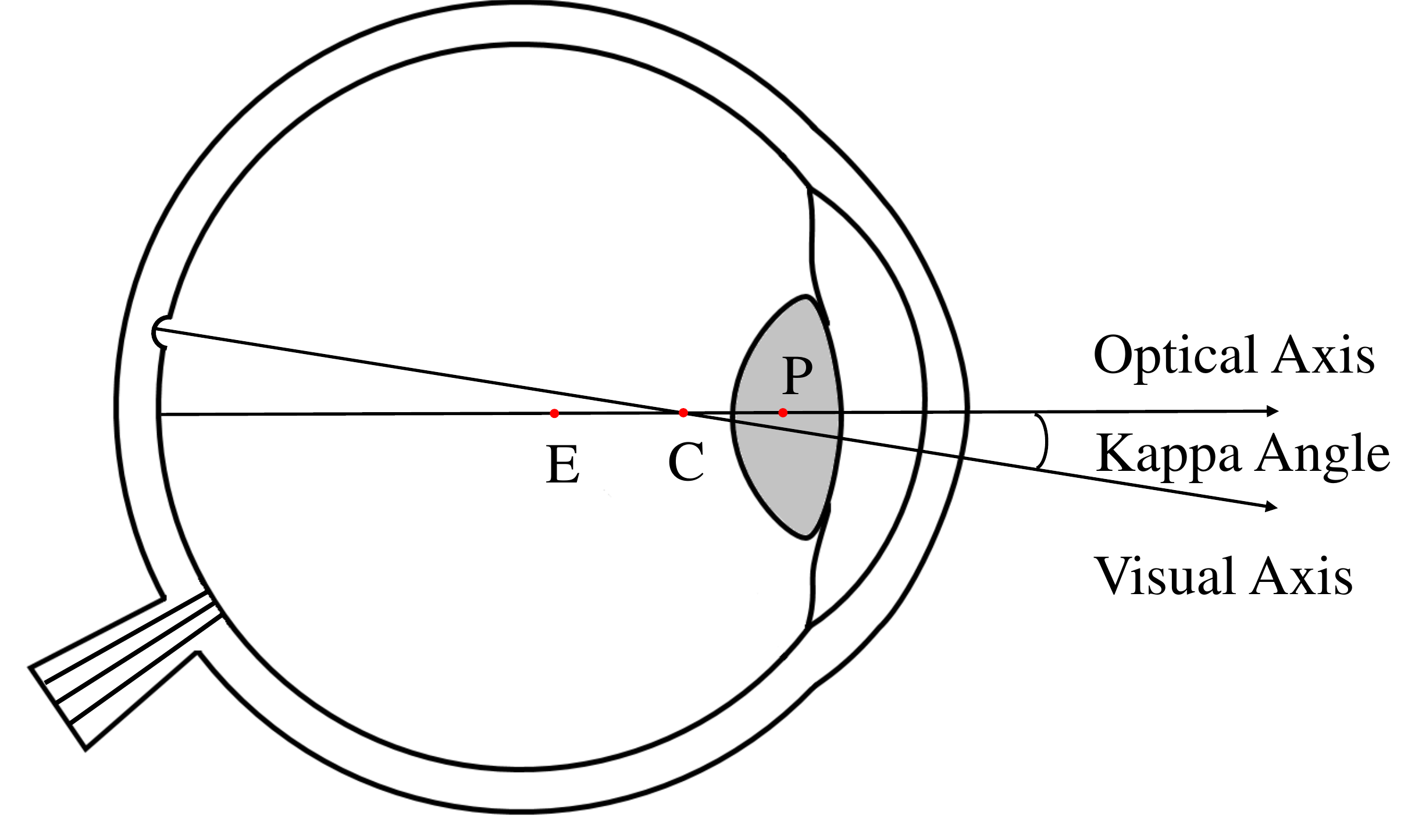}
    \caption{Simplified eyeball model. Optical axis contains pupil center $P$, cornea center $C$ and eyeball rotation center $E$. The visual axis intersects with the optical axis at the cornea center, and the angle between them defines as the kappa angle.}
    \label{fig:eyeball}
\end{figure}

Usually, gaze point tracking includes two branches: the data-driven method and the model drive method. The data-driven method directly maps the eyeball image into the gaze point, while the model-driven method relies on the eyeball model and related optical knowledge. A widely used eye model is the Le Grand model~\cite{le1968light}, shown in Figure \ref{fig:eyeball}. Le Grand model views the cornea as a sphere and ignores the corneal refraction. It is perfect for remote gaze tracking~\cite{hennessey2009improving,coutinho2013improving,morimoto2002detecting} since the approximations in the model do not introduce significant errors in eyeball detection. However, in the near-display device, such an error is non-neglectable. The eyeball in head-mounted gaze tracking must be in the correct position to achieve the ideal visual appearance and angle. False eye position might cause image distortion in VR~\cite{jones2015correction} and object misalignment in MR.

A more precise eyeball model is required in HMD gaze point tracking to mend the modeling error. For example, Nitschke proposed a new eye position estimation method~\cite{nitschke2011image}. Y. Itoh proposed an interaction-free AR calibration method~\cite{itoh2014interaction} for eye position acquisition. However, predefined eyeball parameters introduce deviation among people. Estimating eyeball rotation center with eye poses was discussed in gaze tracking~\cite{swirski2013fully} as well, but the result was highly dependent on pupil contour tracking. In addition, there have been studies of pupil center\cite{fuhl2020neural}  and corneal center\cite{lee2009robust,plopski2015corneal}. A robust eyeball position estimation for various persons  remains a problem in head-mounted systems.

Head-mounted gaze estimation is the basis of gaze-contingent near-eye display realization. A defect of most existing head-mounted gaze tracking methods is the lack of slippage robustness. During common use, a slight relative movement between the head-mounted device and the eyes is unavoidable. Keeping high gaze estimation accuracy with slippage taken into consideration should be realized in practical head-mounted systems. 

Model-based gaze tracking methods~\cite{lukander2013omg} are often considered robust to devise slippage, but complicated system calibration and relatively low accuracy made these methods less famous. The state-of-the-art feature-based gaze estimation method proposed by PupilLab~\cite{kassner2014pupil} achieved a mean angular error lower than $0.6^{\circ}$. A combination of PupilLab eye tracker and VR systems has been studied~\cite{clay2019eye,piumsomboon2017exploring}. However, the method purely relied on the 2D pupil feature and was highly sensitive to slippage. The commonly accepted slippage-robust gaze tracking method Grip~\cite{santini2019get} used a model-based feature (optical axis) and a feature-based gaze mapping model. An acceptable gaze accuracy with slippage robustness was shown. End-to-end learning-based gaze estimation method~\cite{tonsen2017invisibleeye} was proved to be robust and accurate when headset slippage occurred, but model training was data-consuming and required users to collect a large amount of calibration data. 

Recently, slippage in head-mounted gaze tracking has become a critical research problem, and increasing works have been aware of the importance of slippage robustness. Still, few works discussed slippage-robust near-eye display gaze tracking.

\section{System Overview}
\begin{figure}[ht]
    \centering
    \includegraphics[width=\linewidth]{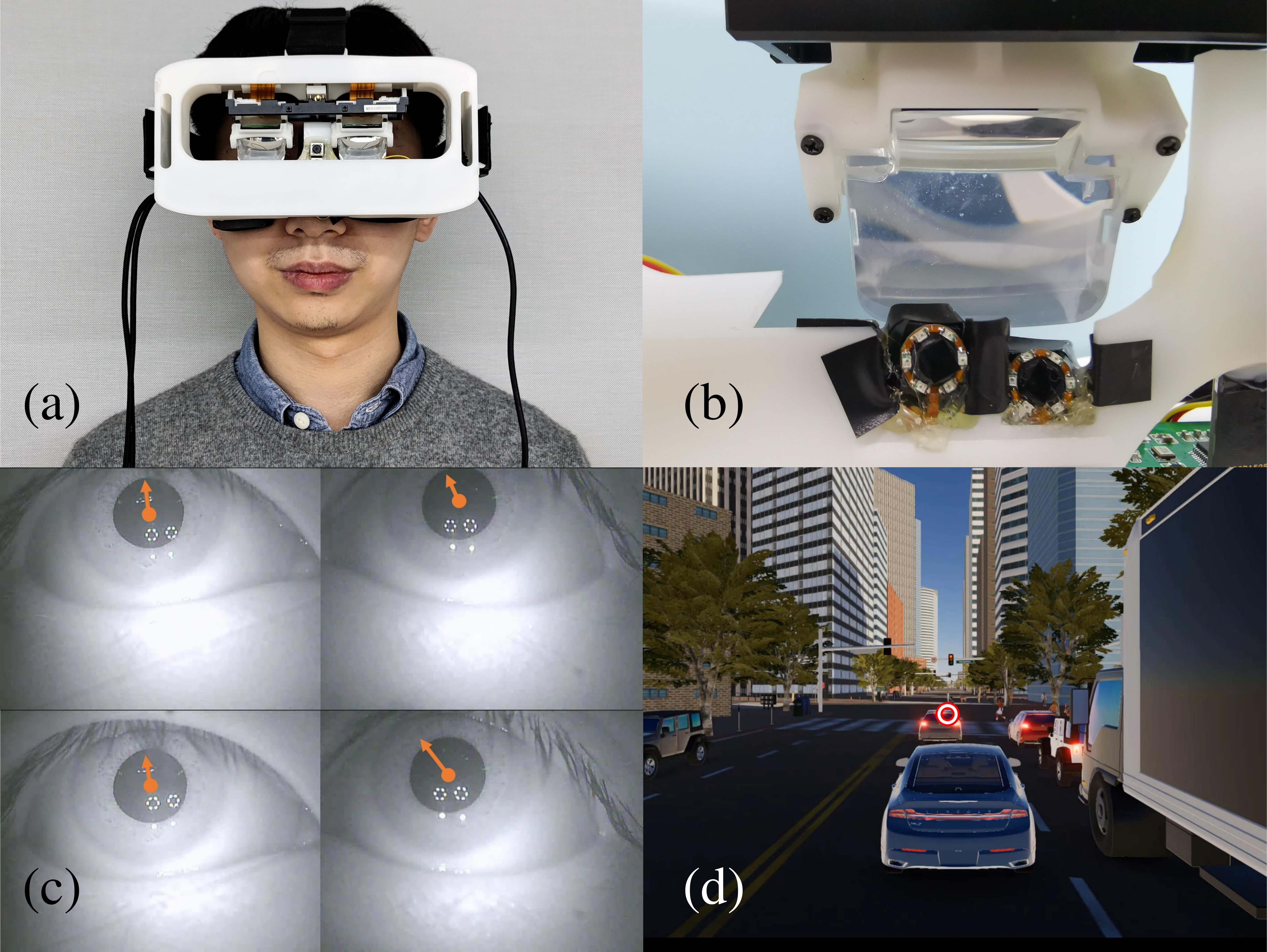}
    \caption{Overview of the proposed system. (a)  Hardware prototype of proposed near-eye display system implemented with an eye tracker. (b) For each eye, there are two near-focus infrared cameras to capture eye images, and six infrared LEDs around each eye camera are used to generate corneal reflection glints. A binocular optical see-through display is implemented in the prototype as a near-eye display. (c)  The subject was asked to gaze at the front vehicle. Pupil centers and optical axes detection results were displayed in eye images. (d) The subject can move and rotate the HMD within a certain range. Results of gaze tracking remained highly accurate. The gaze point estimation result was visualized as red circles in the displayed image. The scene image for gaze estimation demonstration came from svl-simulator (https://www.svlsimulator.com/). }
    \label{fig:setup}
\end{figure}

This section presents our gaze tracking for near-eye display system, where we build hardware devices that can capture eye images, extract features, and have an optical see-through display. Figure \ref{fig:setup} illustrates the workflow of slippage robust gaze tracking for near-eye display.

\subsection{Hardware Setup}

Our headset prototype consists of a near-eye display, 4 eye cameras capturing eye images of $640\times480 pix^2$, and 850 nm infrared LEDs. 

Based on our proposed eye model, determining the eyeball optical axis requires a virtual pupil center and a corneal surface normal vector containing eye camera center. The Virtual pupil centers can be captured in eye images. A Corneal surface normal vector containing eye camera center can be derived by the corneal reflection of a light source coincides with the eye camera center~\cite{nagamatsu2010gaze}. However, implementing a light source inside eye camera is not physically realizable. In our prototype, 6 infrared LEDs around the eye camera are used to generate corneal reflection glints. The centroid of a group of glints approximately provides the reflection position of a light source coincides with the eye camera center. 

A binocular optical see-through display is implemented in the prototype as a near-eye display. Both monocular displays show images of $1920\times1080 pix^2$, and the binocular field of view is $44^{\circ}$. 

\subsection{Feature Detection}

Raw feature detection in eye images includes glint detection and pupil detection. 

Glints appear to be several small bright regions with special patterns in the eye images. A simple thresholding and morphology method will give a centroid result of glints. A problem observed from collected data is that contact lenses might cause glint distortion. The glints reflected by the edge of the contact lens or the unsmooth surface of the contact lens might lead to errors in glint detection. 

\begin{figure}
    \centering
    \subfigure[]{
    \includegraphics[width=0.28\linewidth]{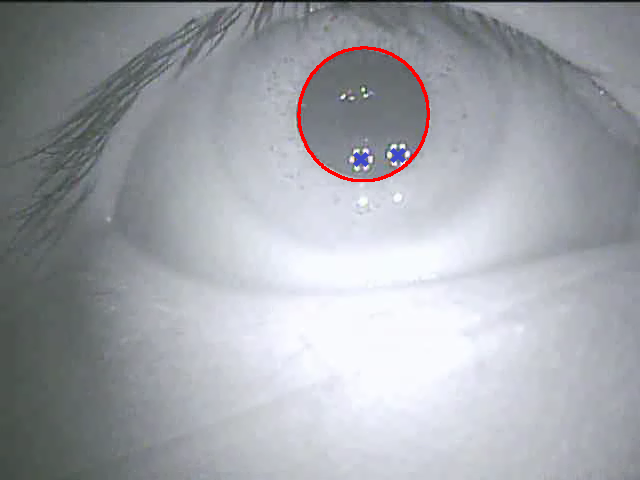}
    }
    \subfigure[]{
    \includegraphics[width=0.28\linewidth]{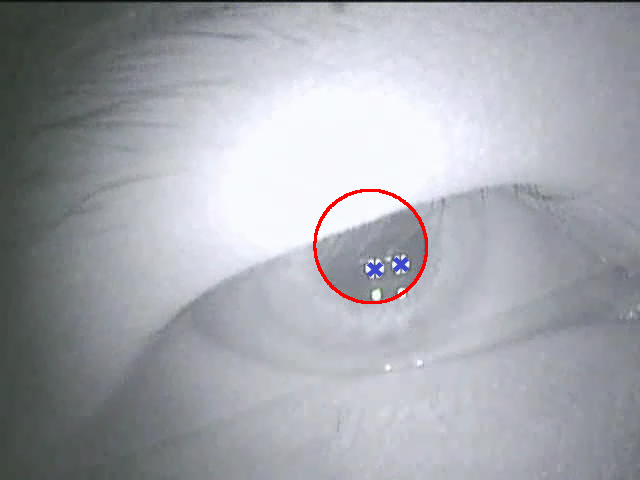}
    }
    \subfigure[]{
    \includegraphics[width=0.28\linewidth]{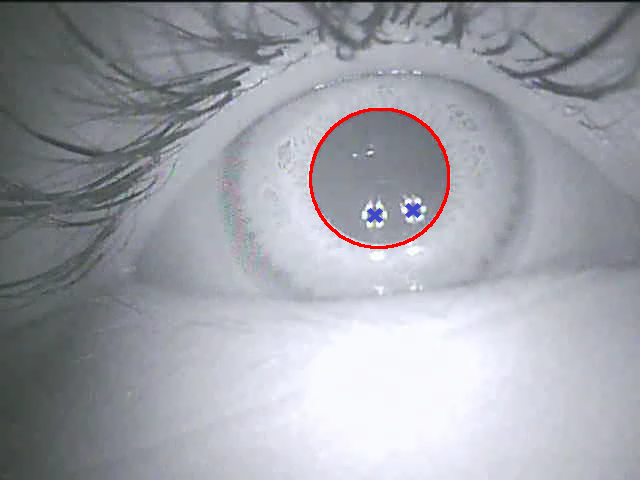}
    }
    \caption{Results of feature detection of (a) complete pupil, (b) pupil partially covered by eyelid, (c) pupil of eye wearing contact len. }
    \label{fig:pupil_det}
\end{figure}

  Pupil detection is a mature technique in gaze tracking technology. We employ radial symmetry transform~\cite{loy2002fast} to detect coarse pupil center, the starburst algorithm~\cite{li2005starburst} to detect pupil edges, PURE~\cite{santini2018pure}  selects pupil edges and fit pupil ellipse. Since eye images are captured in near infrared band, environment illumination produces less effect on image quality and pupil detection remains accurate. Several exemplars of pupil detection results are given in Figure \ref{fig:pupil_det}. 

\section{METHODS}

\subsection{Eyeball Optical Axis Estimation}

Ophthalmology revealed\cite{gatinel2002review} the cornea mean aspheric coefficient $Q = -0.3$, thanks to the naturally prolate asphericity of the cornea reduces spherical aberration by half. Cornea mathematically modeled as
\begin{equation}\label{eq:eyemodel}
    x^2+y^2+(Q+1)*z^2=r^2,
\end{equation}
the z-axis is the axis of rotation and the optical axis of the eye. Our proposed eye model combines the advantages of Nagamatsu's model~\cite{nagamatsu2010gaze} and Lai's model~\cite{lai2014hybrid}, taking aspherical cornea surface and corneal refraction into consideration. 




\begin{figure}[ht]
    \centering
    \includegraphics[width=\linewidth]{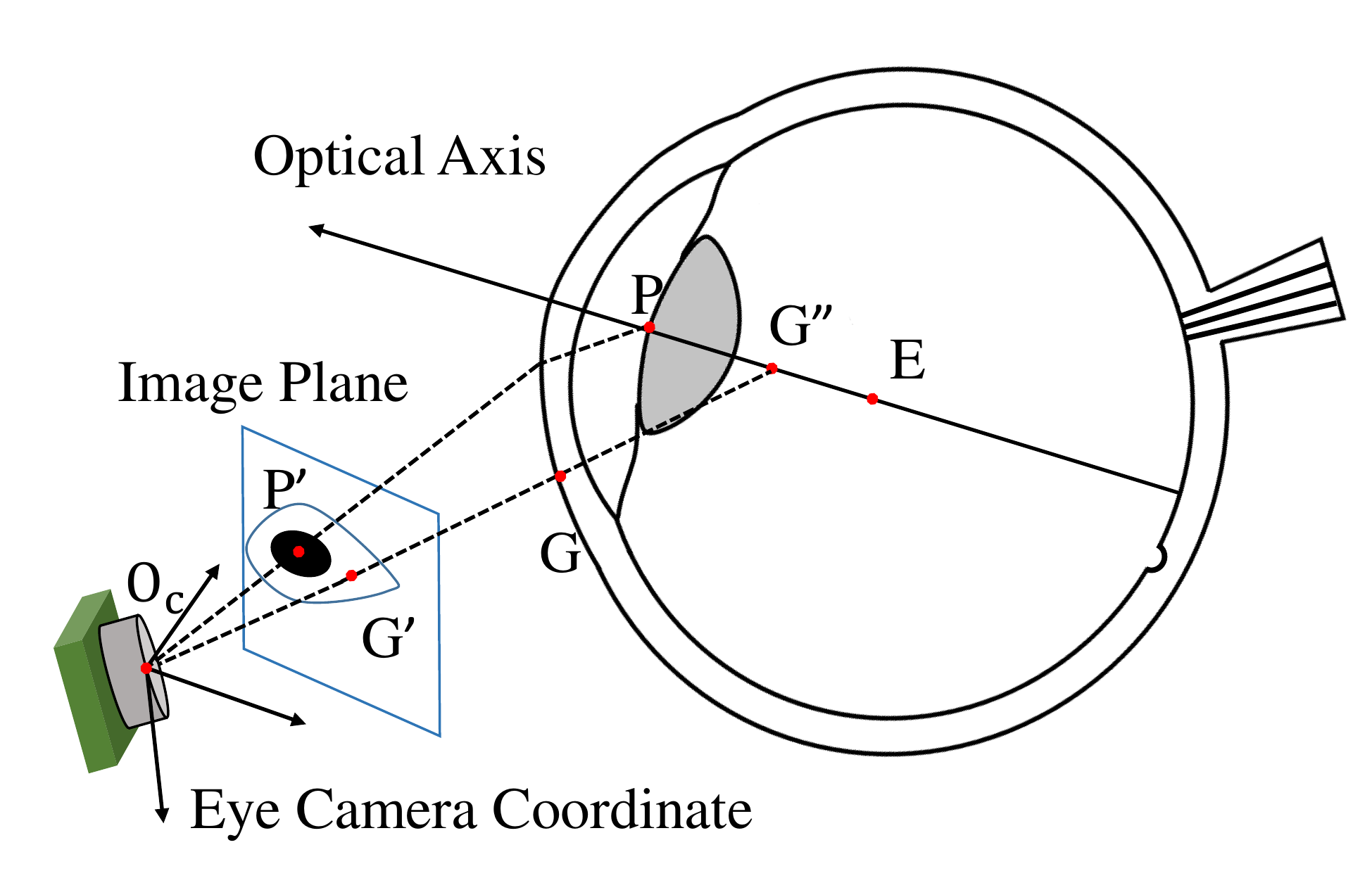}
    \caption{Diagram of proposed eyeball optical axis calculation model. Pupil center $P$ is refracted by cornea and imaged on eye camera image plane at $P'$. Assume an infrared LED coincides with eye camera center $O_{C}$. The corneal reflection $G$ is imaged at $G'$. A plane determined by $O_{C}P'$ and $O_{C}G'$ contains the optical axis.}
    \label{fig:oa_model}
\end{figure}

A diagram of the proposed optical axis calculation model is shown in Figure \ref{fig:oa_model}. Regarding the corneal surface as a rotational ellipsoid, the rotation axis is the eyeball optical axis and thus, normal vector of the corneal surface intersects with eyeball optical axis in 3D space. Considering corneal refraction, the pupil in eye image is a refraction virtual image. According to optical principles, it can be easily proved that the refraction plane contains the virtual pupil center, eye camera center and optical axis~\cite{lai2014hybrid}. Therefore, the plane determined by corneal surface normal vector, eye camera center and virtual pupil center contains eyeball optical axis. Two aforementioned planes intersect at the optical axis. Pupil center and corneal reflection glint in eye image can be detected through aforementioned algorithms. With a calibrated eye camera, 3D line $O_{C}P'$ and $O_{C}G'$ will be obtained. Therefore, the normal vector $ \overrightarrow{\boldsymbol{n}} $ of a plane containing eyeball optical axis is
\begin{equation}
    \overrightarrow{\boldsymbol{n}}={O_{C}P'} \times {O_{C}G'}.
\end{equation}
Furthermore, with stereo eye cameras capturing one eyeball, two planes containing optical axis are obtained and the normal vectors are represented as $ \overrightarrow{\boldsymbol{n}}_1 $ and $ \overrightarrow{\boldsymbol{n}}_2 $ respectively. Then the direction vector of the optical axis is 
\begin{equation}
    \boldsymbol{OA}=\overrightarrow{\boldsymbol{n}}_1 \times \overrightarrow{\boldsymbol{n}}_2.
\end{equation}

\subsection{Eyeball Position Estimation} 
\begin{figure}[ht]
    \centering
    \includegraphics[width=\linewidth]{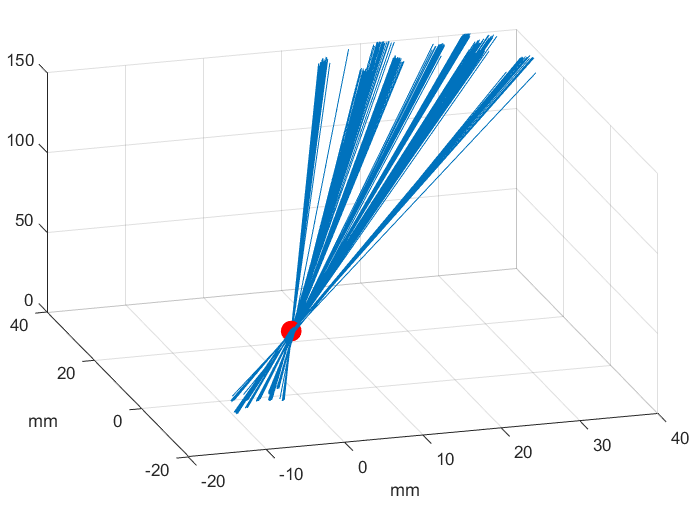}
    \caption{Eyeball optical axis and rotation center. Blue lines indicate the optical axis when subject gazing at calibration markers. The red dot indicates the closest point to all optical axis lines, and is regarded as the estimation of eyeball rotation center.}
    
    \label{fig:eyecenter}
\end{figure}

Since eyeball rotation center is a point on an optical axis and is a fixed point relative to the eye camera if a  device doesn't drift, calculating the intersection point of optical axes of multiple eyeball poses is a way to estimate eyeball rotation center. The distance $d_i$ of eyeball rotation center $E$ to a random optical axis $ OA_i $ is
\begin{equation}
    d_i=\frac{\| EM_i \times OA_i \|}{\| EM_i\|},
\end{equation}
$M_i$ is a random point on line $ OA_i $. The estimation of $E$ is equivalent to the following optimization 
\begin{equation}\label{eq:eyecenter}
    E^\star = \underset{E}{\arg\min} \sum_{i \in \mathcal{I}} {d_i},
\end{equation}
where $\mathcal{I}$ is the index set of all optical axes without slippage and $E^\star$ is the optimal estimation of $E$, the result is shown in Figure \ref{fig:eyecenter}, blue optical axis lines intersect the red center.

\begin{figure}[ht]
    \centering
    \includegraphics[width=0.9\linewidth]{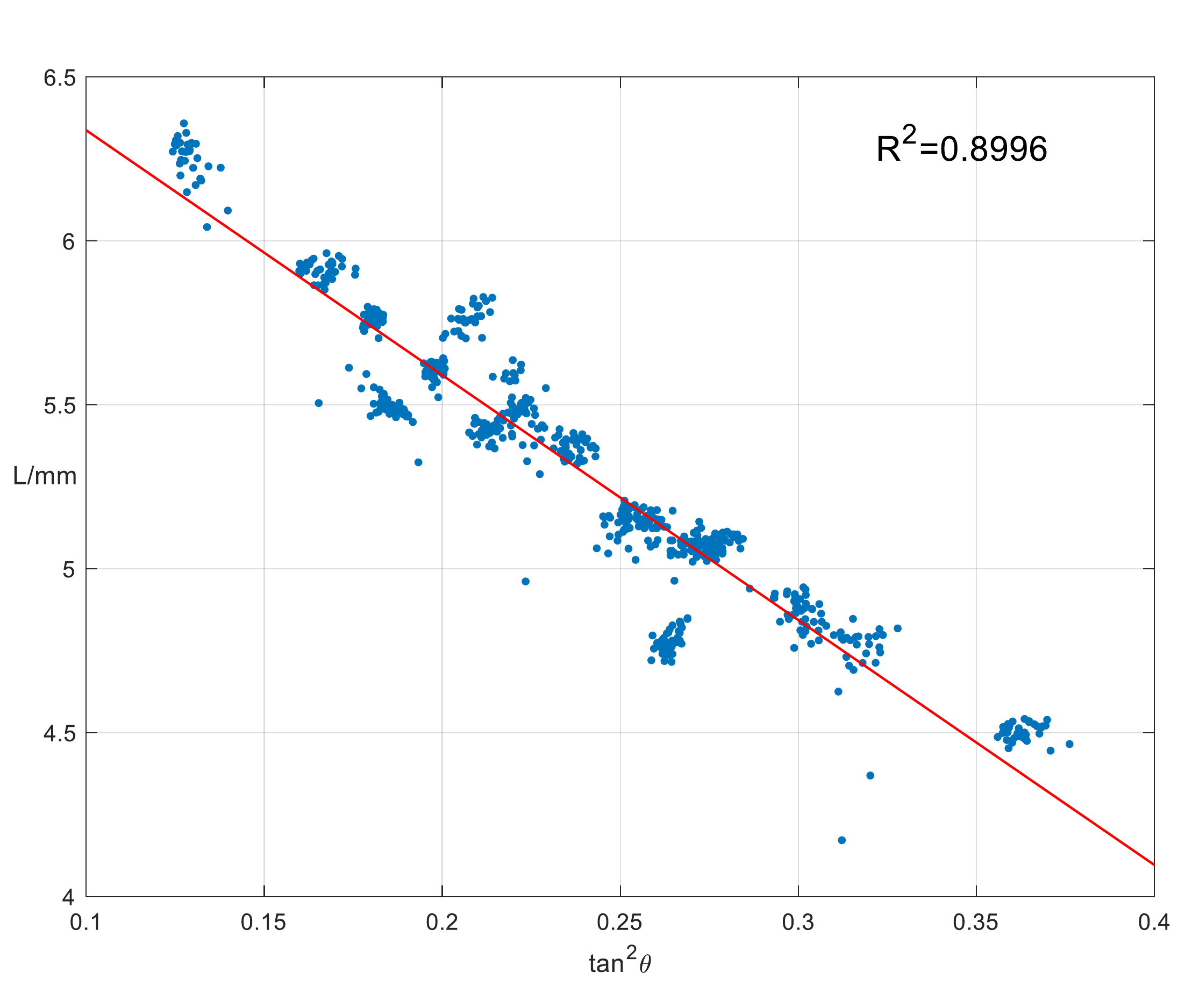}
    
    
    
    \caption{The relationship between $L$ and $\tan^2\theta$. Here $L$ is the distance of $EG''$ shown in figure~\ref{fig:oa_model} and $\theta$ is angle of optical axis and $O_{c}G'$. We could see that $EG''$ and $\tan^2\theta$ are in a linear relationship.}
    \label{fig:eyedistance}
\end{figure}

Apparently, $E^\star$ cannot be obtained with a single optical axis. we only obtain the optical axis direction and the camera glint direction accurately.  The intersection of optical axis and the line of corneal reflection and eye camera center provides a cue. For the mathematical model of corneal asphericity Equation \ref{eq:eyemodel}. In figure~\ref{fig:oa_model} $G''$ is the intersection of optical axis and $O_{c}G'$. The distance $L$ between $E$ and $G''$ is 
\begin{equation}\label{eq:distance_estimation}
    L=t+(1-p)\frac{r}{\sqrt{p}}\frac{1}{\sqrt{p\tan^2\theta+1}},
\end{equation}
 where $\theta$ is the angle of optical axis and $O_{c}G'$, which can be obtained in each frame. t is the distance between the ellipsoidal center of the cornea and the center of rotation of the eyeball. Let $p=Q+1$, so $p,t$ and $r$ are constant value parameters for each individual cornea. Consider Taylor series expansion, Equation \ref{eq:distance_estimation} can be rewritten into a simpler form:
 \begin{equation}\label{eq:distance_estimation_v2}
    L=k_{1}+k_{2}\tan^2\theta.
\end{equation}
 During calibration, $k_{1}$ and $k_{2}$ can be estimated, applying the parameters will produce an estimated eye center with one optical axis with any data frame. The relationship between the experimentally measured distance $L$ and angle is shown in Figure \ref{fig:eyedistance}, the actual data shows the correctness of the method.

\subsection{Slippage-robust Gaze Estimation for Near-eye Display}
The gaze direction is represented by visual axis. With an estimation of optical axis, ask a user to gaze at markers and calibrate the user's kappa angle is a common practice in most eye trackers. To calibrate the difference between optical and visual axes, the 3D transformation between eye camera coordinate and scene coordinate needs to be obtained in advance. Unfortunately, it is a great challenge to calibrate the relation between eye camera and near-eye display system since we cannot plant markers around near-eye display and calibrate like normal screen-based eye tracker, or capture real scene images like a head-mounted scene camera\cite{kasprowski2014guidelines}. Yet, we still can model the near-eye display as a virtual scene camera capturing an image, which is the displaying image. The gaze estimation formula is 
\begin{equation}\label{eq:gaze}
    \begin{bmatrix}
    u\\
    v\\
    1\end{bmatrix}=\lambda\begin{bmatrix}
    f_x & 0 & c_x \\
    0 & f_y & c_y \\
    0 & 0 & 1
    \end{bmatrix}(\boldsymbol{R}(d_{e}VA+T)),
\end{equation}
$\begin{bmatrix} 
u, v\end{bmatrix}^T$ is the gaze point on near-eye display image, $\lambda$ is a normalization coefficient, $f_x, f_y, c_x$, and $c_y$ are intrinsic parameters of the virtual camera, $VA$ is the eyeball visual axis, $d_{e}$ is the distance of virtual image to eyeball,  $E$ is the eyeball rotation center, and $\boldsymbol{R}$ is a transformation from eye camera to virtual camera coordinate. A formula consists of measurable variables, such as $OA$ and $E$, is expected. Based on kappa angle, $VA$ will be replaced by $OA$ and kappa angle, a personal parameter to be calibrated. Since the virtual camera position is related to eye position, translation vector $\boldsymbol{T}$ will be replaced by eyeball position $E$ and the eyeball position during calibration $E_{calib}$. Therefore, Equation \ref{eq:gaze} can be transformed into

\begin{equation}
    \begin{bmatrix}
    u\\
    v\\
    1\end{bmatrix}=\lambda\begin{bmatrix}
    f_x & 0 & c_x \\
    0 & f_y & c_y \\
    0 & 0 & 1
    \end{bmatrix}\boldsymbol{R}(\boldsymbol{R}_{kappa}OA+\frac{1}{d_{e}}(E-E_{calib})),
\end{equation}
$R_{kappa}$ is the rotation matrix produced by kappa angle, and $d_{e}$ is a system-related parameter. During calibration, $\boldsymbol{R}$ and $R_{kappa}$ can be estimated, during a  test, we use eyeball slippage $E-E_{calib}$ to correct gaze estimation. Notice that the intrinsic parameters of the virtual camera can be approximately calculated through the near-eye display field of view parameters and display resolution. The above equation provides the relation between gaze point and monocular optical axis. Since the images shown in near-eye display for both eyes are same in our system, binocular gaze estimation can simply be attained through
\begin{equation}
    \begin{bmatrix}
    u_{bino}\\
    v_{bino}\\\end{bmatrix}=\frac{1}{2}\left(\begin{bmatrix}
    u_{l}\\
    v_{l}\\
    \end{bmatrix}+\begin{bmatrix}
    u_{r}\\
    v_{r}\\
    \end{bmatrix}\right),
\end{equation}
where $\begin{bmatrix} 
u_l, v_l\end{bmatrix}^T$ and $\begin{bmatrix} 
u_r, v_r\end{bmatrix}^T$ are gaze points estimated with left and right eyes optical axes respectively. By averaging two monocular results, the binocular gaze estimation is expected to be more robust and steady. 

\section{EVALUATION}
Limited by differences in hardware and the absence of a consistent data set for our method, the public dataset is unsuitable for the evaluation process. To evaluate the proposed method, we collected data from some different subjects. Experiments were conducted to prove the effectiveness of the eyeball locating method and assess the performance of the gaze tracking method on data collected with headset slippage. Details of the data collection process, experiments, and result analysis are given in this section. 

\subsection{Data Collection}
\begin{figure}[ht]
    \centering
    \includegraphics[width=0.8\linewidth]{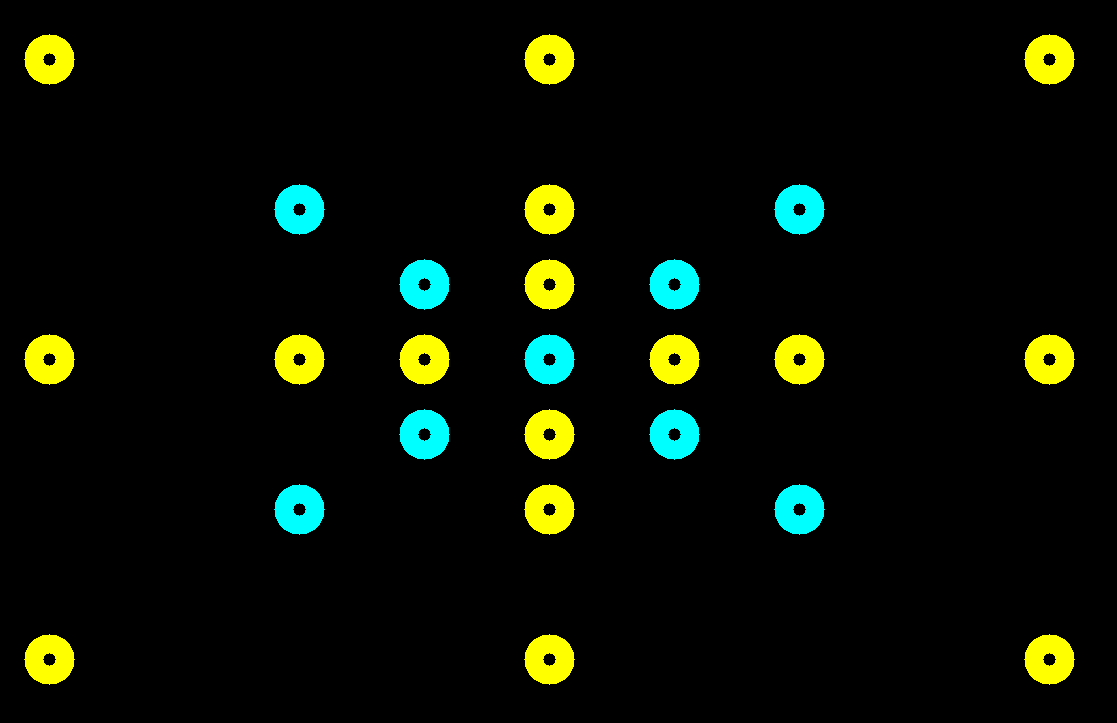}
    \caption{This figure shows the markers used in data collection. Subjects were asked to gaze at the center of each marker on the near-display device to collect calibration, validation, and test data. In calibration recording, data of subjects gazing at nine cyan markers were for calibration, and data of the other 16 markers were for validation. Data of all 25 markers were for testing after subjects remounted the head-mount device. }

    \label{fig:marker}
\end{figure}

In total, nine subjects (four females and five males) between 21 to 27 years old participated in the experiment. All subjects in this experiment had normal or corrected to normal vision. All subjects gave informed consent. The research protocol was approved by the Institutional Review Board at the participating university. 

To ensure all subjects wore the headset properly, they were required not to wear glasses during the collection process, contact lenses were acceptable, and two subjects wore contact lenses during the data collection process. Due to the difference in interpupillary distance (the distance between two pupils), subjects were free to adjust the distance between binocular near-eye displays to attain a larger and clearer field of view. 

This process will collect four recordings for each subject, one for calibration and the other three for testing. The distribution of markers is shown in Figure \ref{fig:marker}. During calibration recording, nine cyan markers are displayed sequentially on the near-eye display, and during test recording, 25 markers are displayed sequentially. Each point was shown for 4 seconds. Subjects were asked to gaze at the center of the shown marker. After each recording, subjects were asked to take off the headset and remount it to simulate slippage in actual use. Calibration and test samples not only produce slippage but also provide some testing points outside the distribution of calibration points.

\begin{table}[t]
    \caption{Mean error and standard deviation of eyeball rotation centers estimation. Eyeball rotation center ground-truth was obtained by optimizing Equation \ref{eq:eyecenter} on testing data, evaluating real-time solve eyeball rotation center performance. }

    
    \label{tab:eyeball}
    \centering
    \begin{tabular}{|c|c|c|c|c|c|c|}
    \hline
    \multirow{3}{*}{Sub.}&\multicolumn{6}{c|}{Eyeball Rotation Center Error/mm}\\
    \cline{2-7}
    ~&\multicolumn{2}{c|}{x}&\multicolumn{2}{c|}{y}&\multicolumn{2}{c|}{z}\\
    \cline{2-7}
    ~&{Mean}&{S.D.}&{Mean}&{S.D.}&{Mean}&{S.D.}\\
    \hline
    1&0.24&0.30&0.26&0.3&0.38&0.33\\
    2&0.18&0.12&0.22&0.19&0.65&0.19\\
    3&0.21&0.18&0.29&0.34&0.76&0.86\\
    4&0.25&0.26&0.47&0.41&1.13&0.73\\
    5&0.34&0.33&0.32&0.36&0.54&0.45\\
    6&0.32&0.41&0.33&0.39&0.95&1.07\\
    7&0.21&0.26&0.35&0.41&0.65&0.69\\
    8&0.14&0.14&0.17&0.18&0.33&0.30\\
    9&0.22&0.24&0.31&0.33&0.56&0.46\\
    \hline
    Mean &0.23&0.25&0.30&0.32&0.66&0.56\\
    \hline
    \end{tabular}
\end{table}

\subsection{Result of Eyeball Position Estimation}

Eyeball position is estimated through the eyeball rotation center. For multiple optical axes without slippage, the eyeball rotation center will be obtained by optimizing Equation \ref{eq:eyecenter}. Figure~\ref{fig:eyecenter} shows the result of eyeball rotation center estimation of one eye through calibration data. When no slippage occurs, the center of the eye can be estimated by the historical data. However, if slippage occurs, a real-time solution of the center is required. We solve the center of the eye in real-time according to Equation \ref{eq:distance_estimation_v2} and use the center calculated by the optimization equation as the ground truth, evaluating real-time solve eyeball rotation center performance. The result of eyeball rotation center estimation is given in Table \ref{tab:eyeball}.
In practice, a stable eyeball location is often much more critical than a highly-accurate location since a fixed error is easily compensated for through calibration. The results show that our method produced a stable and accurate estimation of the eyeball rotation center and thus implies that the previous optical axis results were precise. 


\subsection{Gaze Estimation}

\begin{figure}[ht]
    \centering
    \includegraphics[width=0.9\linewidth]{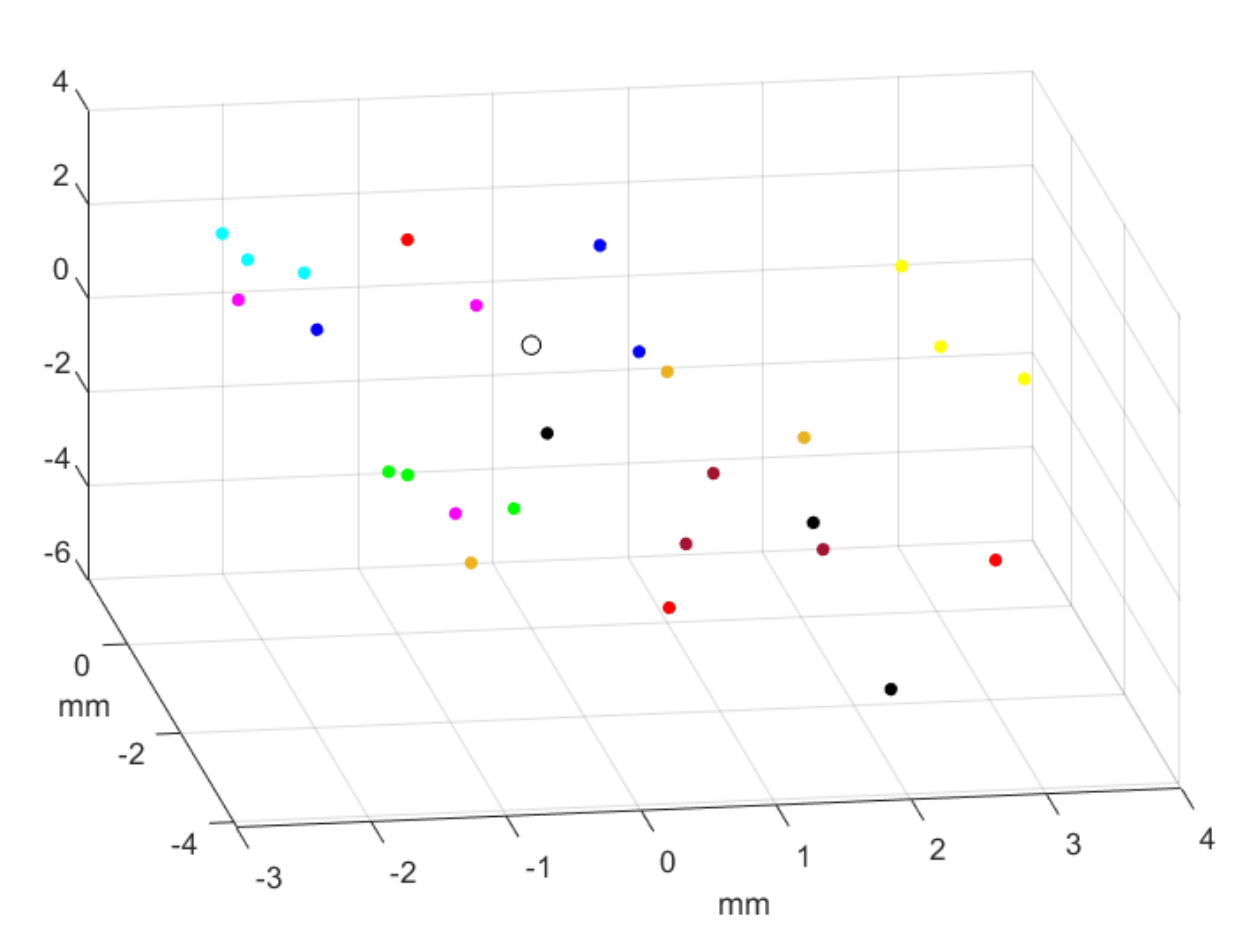}
    \caption{ Slippage distribution after remount. The different colors solid dots represent the different subjects slippage distribution, the hollow point represents the origin. }
    
    \label{fig:eyecenter_distribution}
\end{figure}

The binocular near-eye display implemented in our prototype shows images in two different monocular displays. Therefore, both eyes rotate independently to focus on the same target point in displays and produce a merged view, enabling us to simultaneously realize gaze estimation based on the left or right eye. The results of monocular gaze estimation in the near-eye display of nine subjects are given in Table \ref{tab:mono_result}. Calibration is performed for each subject, using the data of nine markers shown in cyan in Figure \ref{fig:marker}. Data of all 25 points are used for evaluation in the other three recordings, which have been collected after the subject takes off the headset and puts it back on to simulate slippage. In particular, We used the eyeball rotation center found in the previous section to provide the slippage between each test and calibration recording. The result is shown in Figure \ref{fig:eyecenter_distribution}, remount does create slippage.

We compare the performance among Pupil~\cite{kassner2014pupil}, Grip~\cite{santini2019get}, and our method. The mean angular offset of gaze estimation works as a metric to access the results of three methods. Our method outperformed the other two methods on slippage data. Although Pupil achieved a high accuracy after careful calibration, the performance degraded horribly when the headset slipped. Grip maintained a relatively accurate gaze estimation when slippage occurred. However, our method still decreased 27.3\% angular error (32.4\% for the left eye and 22.2\% for the right eye) due to a more reasonable mapping between the optical axis and gaze point. Another fact is that eight testing points lie outside the distribution of calibration points, and the gaze prediction results remain accurate. An extrapolation capability is proven to be possessed by our method. 

\begin{table}[htbp]
    \centering
    \caption{Gaze estimation results of left and right eye. For each subject, calibration was conducted with nine calibration markers data presented in Figure \ref{fig:marker} from one recording. Gaze estimation performance was tested on the other three recordings in which headset slippage was introduced and mean angular offset was presented. Gaze estimation performance comparison was conducted among Pupil\cite{kassner2014pupil}, Grip\cite{santini2019get}, and our method. }
    \label{tab:mono_result}
    \begin{tabular}{|c|c|c|c|c|c|c|}
    \hline
    \multirow{2}{*}{Sub.}&\multicolumn{3}{c|}{\makecell{Left Eye Mean\\Angular Offset/$^\circ$}}&\multicolumn{3}{c|}{\makecell{Right Eye Mean\\Angular Offset/$^\circ$}}\\
    \cline{2-7}
    ~ & Pupil & Grip & Ours & Pupil & Grip & Ours\\
    \hline
    1 & 22.94 & 1.67 & $\boldsymbol{1.27}$ & 27.01 & 1.61 & $\boldsymbol{1.28}$\\
    2 & 19.28 & 1.49 & $\boldsymbol{1.12}$ & 12.68 & 1.26 & $\boldsymbol{1.00}$\\
    3 & 15.51 & $\boldsymbol{1.32}$ & 1.40 & 9.29 & $\boldsymbol{0.91}$ & 0.9\\
    4 & 62.44 & 1.81 & $\boldsymbol{1.07}$ & 42.30 & 1.48 & $\boldsymbol{1.32}$\\
    5 & 12.95 & 2.60 & $\boldsymbol{1.14}$ & 39.64 & 1.25 & $\boldsymbol{0.84}$\\
    6 & 11.51 & 2.06 & $\boldsymbol{1.48}$ & 8.78 & $\boldsymbol{1.09}$ & 1.14\\
    7 & 31.59 & 2.16 & $\boldsymbol{1.00}$ & 41.11 & 2.04 & $\boldsymbol{0.79}$\\
    8 & 11.58 & $\boldsymbol{1.06}$ & 1.14 & 17.20 & 1.10 & $\boldsymbol{0.85}$\\
    9 & 10.33 & 0.84 & $\boldsymbol{0.55}$ & 22.45 & 1.06 & $\boldsymbol{1.02}$\\
    \hline
    Mean & 22.01 & 1.67 & $\boldsymbol{1.13}$ & 24.49 & 1.31 & $\boldsymbol{1.02}$ \\
    \hline
    \end{tabular}
\end{table}

Humans have been using both eyes for a long time, and the brain is more familiar with the visual pattern of both eyes, which is different from that of a single eye. Therefore, binocular information should be more suitable for gaze point estimation. We further tested our methods' binocular gaze estimation results and compared them with state-of-the-art, which is given in Table \ref{tab:bino_result}. The binocular gaze estimation is obtained through a simple averaging of both monocular gaze results. The results show that the mean angular error decreased 25.5\% in contrast to the monocular results of our method, which improved much more than Grip. At the same time, our method outperforms the grip method by 100\%, and the angular offset decreases from $1.51^\circ$ to $0.76^\circ$. The decrease in mean angular error implies an improvement in slippage robustness. 

Experimental results show that based on the proposed eye model, our method achieved high accuracy and high slippage robustness in near-eye display gaze estimation. Extrapolation enabled users to calibrate with a small number of markers lying in a limited field of view while realizing gaze tracking in a larger space. 

\begin{table}[htbp]
    \centering
    \caption{Binocular gaze estimation results. Comparison among state-of-the-art methods and our method show that the proposed method exhibit a more accurate and slippage-robust performance in near-eye display gaze estimation. An improvement in gaze accuracy is brought by binocular information, implying that slippage robustness will be higher if binocular data is utilized. }
    \label{tab:bino_result}
    \begin{tabular}{|c|m{1.5cm}<{\centering}|m{1.5cm}<{\centering}|m{1.5cm}<{\centering}|}
    \hline
    \multirow{2}{*}{Sub.}&\multicolumn{3}{c|}{Binocular Mean Angular Offset/$^\circ$}\\
    \cline{2-4}
    ~ & Pupil & Grip & Ours\\
    \hline
    1&28.60&2.11&$\boldsymbol{0.81}$\\
    2&16.36&1.46&$\boldsymbol{0.87}$\\
    3&33.74&2.12&$\boldsymbol{0.84}$\\
    4&66.28&1.48&$\boldsymbol{0.66}$\\
    5&60.83&1.19&$\boldsymbol{0.75}$\\
    6&11.95&1.59&$\boldsymbol{0.99}$\\
    7&55.05&1.69&$\boldsymbol{0.65}$\\
    8&15.31&1.06&$\boldsymbol{0.75}$\\
    9&19.75&0.91&$\boldsymbol{0.53}$\\
    \hline
    Mean&34.21&1.51&$\boldsymbol{0.76}$\\
    \hline
    \end{tabular}
\end{table}

\section{CONCLUSION}
In this paper, we proposed an eyeball optical axis and position estimation method based on the aspheric eye model. Furthermore, a slippage-robust gaze tracking system was deployed based on the relation between gaze point, eyeball's optical axis, and eyeball's position. Experimental results show that the proposed method maintained high-accuracy gaze estimation when headset slippage occurred. Slippage robustness is crucial for a practical gaze tracking system, and our method guarantees a promising future for the near-eye display headset. 







\bibliographystyle{ieeetr}
\bibliography{root}

\begin{thebibliography}{10}

\bibitem{liu2019high}
H.~Liu, Z.~Zhang, X.~Xie, Y.~Zhu, Y.~Liu, Y.~Wang, and S.-C. Zhu,
  ``High-fidelity grasping in virtual reality using a glove-based system,'' in
  {\em 2019 international conference on robotics and automation (icra)},
  pp.~5180--5186, IEEE, 2019.

\bibitem{lin2022comparison}
T.-C. Lin, A.~U. Krishnan, and Z.~Li, ``Comparison of haptic and augmented
  reality visual cues for assisting tele-manipulation,'' in {\em 2022
  International Conference on Robotics and Automation (ICRA)}, pp.~9309--9316,
  IEEE, 2022.

\bibitem{patney2016towards}
A.~Patney, M.~Salvi, J.~Kim, A.~Kaplanyan, C.~Wyman, N.~Benty, D.~Luebke, and
  A.~Lefohn, ``Towards foveated rendering for gaze-tracked virtual reality,''
  {\em ACM Transactions on Graphics (TOG)}, vol.~35, no.~6, pp.~1--12, 2016.

\bibitem{konrad2020gaze}
R.~Konrad, A.~Angelopoulos, and G.~Wetzstein, ``Gaze-contingent ocular parallax
  rendering for virtual reality,'' {\em ACM Transactions on Graphics (TOG)},
  vol.~39, no.~2, pp.~1--12, 2020.

\bibitem{krajancich2020optimizing}
B.~Krajancich, P.~Kellnhofer, and G.~Wetzstein, ``Optimizing depth perception
  in virtual and augmented reality through gaze-contingent stereo rendering,''
  {\em ACM Transactions on Graphics (TOG)}, vol.~39, no.~6, pp.~1--10, 2020.

\bibitem{zheng2018rapid}
C.~Zheng and T.~Usagawa, ``A rapid webcam-based eye tracking method for human
  computer interaction,'' in {\em 2018 International Conference on Control,
  Automation and Information Sciences (ICCAIS)}, pp.~133--136, IEEE, 2018.

\bibitem{majaranta2019eye}
P.~Majaranta, K.-J. R{\"a}ih{\"a}, A.~Hyrskykari, and O.~{\v{S}}pakov, ``Eye
  movements and human-computer interaction,'' in {\em Eye Movement Research},
  pp.~971--1015, Springer, 2019.

\bibitem{saran2018human}
A.~Saran, S.~Majumdar, E.~S. Short, A.~Thomaz, and S.~Niekum, ``Human gaze
  following for human-robot interaction,'' in {\em 2018 IEEE/RSJ International
  Conference on Intelligent Robots and Systems (IROS)}, pp.~8615--8621, IEEE,
  2018.

\bibitem{blattgerste2018advantages}
J.~Blattgerste, P.~Renner, and T.~Pfeiffer, ``Advantages of eye-gaze over
  head-gaze-based selection in virtual and augmented reality under varying
  field of views,'' in {\em Proceedings of the Workshop on Communication by
  Gaze Interaction}, pp.~1--9, 2018.

\bibitem{li2020optical}
R.~Li, E.~Whitmire, M.~Stengel, B.~Boudaoud, J.~Kautz, D.~Luebke, S.~Patel, and
  K.~Ak{\c{s}}it, ``Optical gaze tracking with spatially-sparse single-pixel
  detectors,'' in {\em 2020 IEEE International Symposium on Mixed and Augmented
  Reality (ISMAR)}, pp.~117--126, IEEE, 2020.

\bibitem{clay2019eye}
V.~Clay, P.~K{\"o}nig, and S.~Koenig, ``Eye tracking in virtual reality,'' {\em
  Journal of Eye Movement Research}, vol.~12, no.~1, 2019.

\bibitem{niehorster2020impact}
D.~C. Niehorster, T.~Santini, R.~S. Hessels, I.~T. Hooge, E.~Kasneci, and
  M.~Nystr{\"o}m, ``The impact of slippage on the data quality of head-worn eye
  trackers,'' {\em Behavior Research Methods}, vol.~52, no.~3, pp.~1140--1160,
  2020.

\bibitem{santini2019get}
T.~Santini, D.~C. Niehorster, and E.~Kasneci, ``Get a grip: Slippage-robust and
  glint-free gaze estimation for real-time pervasive head-mounted eye
  tracking,'' in {\em Proceedings of the 11th ACM symposium on eye tracking
  research \& applications}, pp.~1--10, 2019.

\bibitem{le1968light}
Y.~Le~Grand, {\em Light, colour and vision}.
\newblock Chapman \& Hall, 1968.

\bibitem{hennessey2009improving}
C.~A. Hennessey and P.~D. Lawrence, ``Improving the accuracy and reliability of
  remote system-calibration-free eye-gaze tracking,'' {\em IEEE transactions on
  biomedical engineering}, vol.~56, no.~7, pp.~1891--1900, 2009.

\bibitem{coutinho2013improving}
F.~L. Coutinho and C.~H. Morimoto, ``Improving head movement tolerance of
  cross-ratio based eye trackers,'' {\em International journal of computer
  vision}, vol.~101, no.~3, pp.~459--481, 2013.

\bibitem{morimoto2002detecting}
C.~H. Morimoto, A.~Amir, and M.~Flickner, ``Detecting eye position and gaze
  from a single camera and 2 light sources,'' in {\em Object recognition
  supported by user interaction for service robots}, vol.~4, pp.~314--317,
  IEEE, 2002.

\bibitem{jones2015correction}
J.~A. Jones, L.~C. Dukes, D.~M. Krum, M.~T. Bolas, and L.~F. Hodges,
  ``Correction of geometric distortions and the impact of eye position in
  virtual reality displays,'' in {\em 2015 International Conference on
  Collaboration Technologies and Systems (CTS)}, pp.~77--83, IEEE, 2015.

\bibitem{nitschke2011image}
C.~Nitschke, ``Image-based eye pose and reflection analysis for advanced
  interaction techniques and scene understanding,'' 2011.

\bibitem{itoh2014interaction}
Y.~Itoh and G.~Klinker, ``Interaction-free calibration for optical see-through
  head-mounted displays based on 3d eye localization,'' in {\em 2014 IEEE
  symposium on 3d user interfaces (3dui)}, pp.~75--82, IEEE, 2014.

\bibitem{swirski2013fully}
L.~Swirski and N.~Dodgson, ``A fully-automatic, temporal approach to single
  camera, glint-free 3d eye model fitting,'' {\em Proc. PETMEI}, pp.~1--11,
  2013.

\bibitem{fuhl2020neural}
W.~Fuhl, H.~Gao, and E.~Kasneci, ``Neural networks for optical vector and eye
  ball parameter estimation,'' in {\em ACM Symposium on Eye Tracking Research
  and Applications}, pp.~1--5, 2020.

\bibitem{lee2009robust}
E.~C. Lee and K.~R. Park, ``A robust eye gaze tracking method based on a
  virtual eyeball model,'' {\em Machine Vision and Applications}, vol.~20,
  no.~5, pp.~319--337, 2009.

\bibitem{plopski2015corneal}
A.~Plopski, Y.~Itoh, C.~Nitschke, K.~Kiyokawa, G.~Klinker, and H.~Takemura,
  ``Corneal-imaging calibration for optical see-through head-mounted
  displays,'' {\em IEEE transactions on visualization and computer graphics},
  vol.~21, no.~4, pp.~481--490, 2015.

\bibitem{lukander2013omg}
K.~Lukander, S.~Jagadeesan, H.~Chi, and K.~M{\"u}ller, ``Omg! a new robust,
  wearable and affordable open source mobile gaze tracker,'' in {\em
  Proceedings of the 15th international conference on Human-computer
  interaction with mobile devices and services}, pp.~408--411, 2013.

\bibitem{kassner2014pupil}
M.~Kassner, W.~Patera, and A.~Bulling, ``Pupil: an open source platform for
  pervasive eye tracking and mobile gaze-based interaction,'' in {\em
  Proceedings of the 2014 ACM international joint conference on pervasive and
  ubiquitous computing: Adjunct publication}, pp.~1151--1160, 2014.

\bibitem{piumsomboon2017exploring}
T.~Piumsomboon, G.~Lee, R.~W. Lindeman, and M.~Billinghurst, ``Exploring
  natural eye-gaze-based interaction for immersive virtual reality,'' in {\em
  2017 IEEE symposium on 3D user interfaces (3DUI)}, pp.~36--39, IEEE, 2017.

\bibitem{tonsen2017invisibleeye}
M.~Tonsen, J.~Steil, Y.~Sugano, and A.~Bulling, ``Invisibleeye: Mobile eye
  tracking using multiple low-resolution cameras and learning-based gaze
  estimation,'' {\em Proceedings of the ACM on Interactive, Mobile, Wearable
  and Ubiquitous Technologies}, vol.~1, no.~3, pp.~1--21, 2017.

\bibitem{nagamatsu2010gaze}
T.~Nagamatsu, Y.~Iwamoto, J.~Kamahara, N.~Tanaka, and M.~Yamamoto, ``Gaze
  estimation method based on an aspherical model of the cornea: surface of
  revolution about the optical axis of the eye,'' in {\em Proceedings of the
  2010 Symposium on Eye-Tracking Research \& Applications}, pp.~255--258, 2010.

\bibitem{loy2002fast}
G.~Loy and A.~Zelinsky, ``A fast radial symmetry transform for detecting points
  of interest,'' in {\em European Conference on Computer Vision}, pp.~358--368,
  Springer, 2002.

\bibitem{li2005starburst}
D.~Li, D.~Winfield, and D.~J. Parkhurst, ``Starburst: A hybrid algorithm for
  video-based eye tracking combining feature-based and model-based
  approaches,'' in {\em 2005 IEEE Computer Society Conference on Computer
  Vision and Pattern Recognition (CVPR'05)-Workshops}, pp.~79--79, IEEE, 2005.

\bibitem{santini2018pure}
T.~Santini, W.~Fuhl, and E.~Kasneci, ``Pure: Robust pupil detection for
  real-time pervasive eye tracking,'' {\em Computer Vision and Image
  Understanding}, vol.~170, pp.~40--50, 2018.

\bibitem{gatinel2002review}
D.~Gatinel, M.~Haouat, and T.~Hoang-Xuan, ``A review of mathematical
  descriptors of corneal asphericity,'' {\em Journal Francais d'ophtalmologie},
  vol.~25, no.~1, pp.~81--90, 2002.

\bibitem{lai2014hybrid}
C.-C. Lai, S.-W. Shih, and Y.-P. Hung, ``Hybrid method for 3-d gaze tracking
  using glint and contour features,'' {\em IEEE Transactions on Circuits and
  Systems for Video Technology}, vol.~25, no.~1, pp.~24--37, 2014.

\bibitem{kasprowski2014guidelines}
P.~Kasprowski, K.~Har{\k{e}}{\.z}lak, and M.~Stasch, ``Guidelines for the eye
  tracker calibration using points of regard,'' in {\em Information
  Technologies in Biomedicine, Volume 4}, pp.~225--236, Springer, 2014.

\end{thebibliography}

\end{document}